\pdfoutput=1
\documentclass[11pt]{article}

\usepackage{amsmath, amsfonts, amssymb}
\usepackage{textcomp}
\usepackage{stfloats}
\usepackage{url}
\usepackage{verbatim}
\usepackage{graphicx}
\usepackage{latexsym}
\usepackage[hidelinks]{hyperref}
\usepackage{gensymb}
\usepackage{mathtools}
\usepackage{comment}
\usepackage{multirow}
\usepackage{adjustbox}
\usepackage{rotating}
\usepackage{caption}
\usepackage{tabularx}
\usepackage{booktabs}
\usepackage{subcaption}
\usepackage{enumitem} % enumerate with custom prefix
\usepackage[ruled,linesnumbered]{algorithm2e}
\usepackage[dvipsnames]{xcolor}
\usepackage{float}

\usepackage{array} 
\usepackage{pdflscape}
\usepackage{supertabular, longtable}
\usepackage{colortbl}

\usepackage{acl}
\AtBeginDocument{\nolinenumbers}  % Disable line numbers for arXiv while keeping anonymous

\usepackage{times}
\usepackage[T1]{fontenc}
\usepackage[utf8]{inputenc}
\usepackage{listings}
\newcommand{\prompthead}[1]{%
    \par\noindent%
    \lstinline[basicstyle=\bfseries\small\ttfamily, breaklines=true]|#1|%
    \par\vspace{2pt}%
}

\usepackage{microtype}

\usepackage{inconsolata}

\title{MiRAGE: A Multiagent Framework for Generating Multimodal Multihop Question-Answer Dataset for RAG Evaluation}
\author{
  \textbf{Chandan Kumar Sahu\textsuperscript{1}}\thanks{Equal contribution.},
  \textbf{Premith Kumar Chilukuri\textsuperscript{1}}\footnotemark[1],
  \textbf{Matthew Hetrich\textsuperscript{1}}\footnotemark[1]\thanks{Corresponding author: \href{mailto:matthew.hetrich1@us.abb.com}{matthew.hetrich1@us.abb.com}}
\\
\\
  \textsuperscript{1}ABB Inc
}

\begin{document}
\maketitle
\begin{abstract}
The rapid evolution of Retrieval-Augmented Generation (RAG) toward multimodal, high-stakes enterprise applications has outpaced the development of domain specific evaluation benchmarks. 
Existing datasets often rely on general-domain corpora or purely textual retrieval, failing to capture the complexity of specialized technical documents where information is inextricably multimodal and reasoning requires synthesizing disjoint evidence. 
We address this gap by introduing \textbf{MiRAGE}\footnote{\url{https://github.com/ChandanKSahu/MiRAGE}}, a \textbf{M}ultiagent framework for \textbf{RAG} systems \textbf{E}valuation, that leverages a collaborative swarm of specialized agents to generate verified, domain-specific, multimodal, and multi-hop Question-Answer datasets. 
MiRAGE orchestrates a swarm of specialized agents: a recursive context optimization loop to aggregate scattered evidence, an adversarial verifier agent to guarantee factual grounding, and an agent to recognize the expert persona and the relevant domain to mimic expert cognitive workflows. 
Extensive empirical evaluation across four distinct domains (regulations, finance, quantitative biology, and journalism) demonstrates that MiRAGE generates datasets with significantly higher reasoning complexity ($>2.3$ average hops) and factual faithfulness.
Our ablation studies point that MiRAGE can be powered by LLMs if textual descriptions of the images are available.
Visual grounding still remains a frontier.
By automating the creation of gold standard evaluation datasets that reflect the latent thematic structure of proprietary corpora, MiRAGE provides the necessary infrastructure to rigorously benchmark the next generation information retrieval systems.
\end{abstract}

\section{Introduction}

Large Language Models (LLMs) have demonstrated remarkable capabilities in encoding world knowledge within their parameters. However, they face fundamental limitations regarding rare entities, cutoff dates, and the high computational cost of retraining to update information \citep{kandpal2023large, mallen2023not}.
Retrieval-Augmented Generation (RAG) has emerged as the de facto solution to these challenges, mitigating hallucinations by grounding generation in external, non-parametric knowledge bases \citep{lewis2020retrieval}.
As RAG systems are increasingly deployed in high-stakes enterprise domains ranging from medical diagnosis \cite{xiong2024benchmarking} to wind energy \cite{meyur2025weqa}, the imperative for rigorous, scenario-specific evaluation has intensified.

Despite the rapid adoption of RAG, evaluating these systems remains a non-trivial challenge.
Standard benchmarks, such as Natural Questions \citep{kwiatkowski2019natural} or MS MARCO \citep{nguyen2016ms}, typically rely on open-domain data that fails to reflect the complexity of specialized corpora.
% \citep{zheng2025revolutionizing}.
In real-world environments, knowledge is rarely confined to text.
It is inextricably multimodal, locked within charts, technical diagrams, and complex layouts.
Recent work on Multimodal RAG, such as MuRAG \citep{chen2022murag}, highlights that models restricted to textual retrieval neglect the massive amount of knowledge present in visual modalities.
Furthermore, while single-hop retrieval is relatively mature, existing systems struggle significantly with multi-hop queries that require synthesizing disjoint pieces of evidence scattered within a document \citep{tang2024multihop}.
% , yang2018hotpotqa}. 

To bridge this gap, research has pivoted toward synthetic dataset generation, as obtaining human-annotated data for multi-hop reasoning is both time-consuming and resource-intensive \citep{wu2024synthetic}.
However, current synthetic frameworks often suffer from critical deficiencies.
Pipelines like DataMorgana \citep{filice2025generating} or SMMQG \citep{wu2024synthetic} generally employ linear generation strategies that lack robust feedback mechanisms.
This often results in hallucinated evaluation datasets.
% , where the generated ground truth is not supported by the retrieved context, or datasets that lack the reasoning depth required to stress-test modern reasoning models \citep{es2024ragas}.
There is a paucity of high-quality benchmarks that simultaneously address domain specificity, multimodality, and complex reasoning steps.

To address these limitations, we introduce \textbf{MiRAGE}, a Multi-Agentic framework designed to generate robust multihop multimodal expert-level evaluation datasets.
Unlike linear prompting strategies, MiRAGE orchestrates a swarm of specialized agents to mimic the cognitive workflow of a domain expert.
We propose a dynamic \textit{context optimization loop}, where a retrieval agent recursively builds a semantic context window, gathering scattered evidence to support complex inquiries before a question-answer pair is formulated.
Crucially, we address the reliability of synthetic data through an adversarial verification phase, employing a dedicated agent to fact-check generated answers against source context ensuring that the generated insight is consistent with the source. 

Our contributions are summarized as follows:
\begin{enumerate}
    \item We propose a model-agnostic, multi-agent framework that automates the ingestion and semantic segmentation of complex multimodal documents, preserving the semantic dependencies between text and visual elements.
    % often lost in standard OCR processes.
    \item We introduce a novel generative methodology that utilizes recursive context expansion and persona injection to produce multi-hop QA pairs. This allows for the creation of questions that require logical deduction across disjoint chunks, surpassing the complexity of extractive QA.
    \item We provide an extensive empirical evaluation across four distinct domains: regulations, finance, science, and journalism.
    We demonstrate that MiRAGE generates datasets with significantly higher reasoning complexity while strictly adhering to the latent thematic distribution of the source domain.
    \item Our ablation study revealed that the domain/persona injection, multihop context and QA verification play a crucial role in the quality of the generated QA dataset. MiRAGE can be powered by LLMs if the textual descriptions of the images contained in the document are available.
\end{enumerate}

\section{Literature Review}

% The rapid adoption of Retrieval-Augmented Generation (RAG) in enterprise settings has necessitated rigorous evaluation frameworks to ensure reliability and accuracy. While traditional benchmarks have served the community well, the shift toward proprietary, multimodal, and complex data sources has exposed significant gaps in existing evaluation methodologies. This section reviews current approaches in RAG evaluation, multimodal integration, and multi-hop reasoning, highlighting the limitations that necessitate the proposed \textbf{MiRAGE} framework.

\subsection{Evaluation of RAG Systems}

Evaluating RAG systems remains a non-trivial challenge due to the dual dependencies of retrieval precision and generation faithfulness. 
Traditional evaluation often relies on static, general-domain datasets such as Natural Questions \cite{kwiatkowski2019natural}, MS MARCO \cite{nguyen2016ms} and MIRAGE \cite{park2025mirage}. 
However, these benchmarks fail to capture the nuances of domain-specific corpora. 
Consequently, the generation of synthetic datasets became inevitable. 
Frameworks like \textit{DataMorgana} \citep{filice2025generating} and \textit{DQABench} \citep{zheng2025revolutionizing} leverage LLMs to generate QA pairs for metrics-based evaluation \citep{es2024ragas}.
% Frameworks like \textit{DataMorgana} \citep{filice2025generating} and \textit{DQABench} \citep{zheng2025revolutionizing} leverage LLMs to generate QA pairs to evaluate systems on metrics such as faithfulness, answer relevance, and context relevance \citep{es2024ragas}.

The proprietary nature of \textit{DataMorgana} and the schema-based nature of \textit{DQABench}, limit their usage and generalizability.
% Moreover, current synthetic generation methods often suffer from content hallucination.
% where the generated ground truth does not align perfectly with the source text or fails to reflect the complexity of real-world queries.
Frameworks like \textit{RAGProbe} \citep{sivasothy2025ragprobe} highlight that system failures occur most frequently when prompts combine multiple constraints. 
Thus, most synthetic generators produce simple, single-constraint queries. 
MiRAGE addresses the reliability of synthetic data through a dedicated QA generation agent with a distinct verifier agent that fact-checks generated answers to mitigate hallucination.

\subsection{Multimodal Context}

Real-world documents (E.g., technical standards, manuals, and research papers) are inherently multimodal, interleaving text with tables, charts, and diagrams.
Recent literature on Visual QA \citep{kim2025visual} and Multimodal RAG \citep{mei2025survey} emphasizes that critical information is often locked in non-textual formats.
Recent works like \textit{SPIQA} \citep{pramanick2024spiqa} and \textit{WikiMixQA} \citep{foroutan2025wikimixqa} have attempted to bridge the multimodal gap by interpreting scientific figures.
\textit{MuRAG} \citep{chen2022murag} requires joint reasoning over images and text.
However, the critical bottleneck of a unified perception stage still persists \citep{zhao2023retrieving}. 
Modalities are often processed in isolation, breaking the document's semantic flow.
The standard OCR frequently degrades tabular structure. 
MiRAGE addresses this by implementing a multimodal ingestion phase where a vision agent generates descriptions and extracts table structures into an enriched markdown. 
This preserves the semantic proximity between textual and visual elements in a shared vector space, mirroring the source layout.

% Real-world documents (E.g., technical standards, manuals, and research papers) are inherently multimodal, interleaving text with tables, charts, and diagrams.
% Recent literature on Visual QA \citep{kim2025visual} and Multimodal RAG \citep{mei2025survey} emphasizes that critical information is often locked in non-textual formats.
% Datasets like \textit{SPIQA} \citep{pramanick2024spiqa} and \textit{WikiMixQA} \citep{foroutan2025wikimixqa} have attempted to bridge this gap by focusing on interpreting scientific figures and tables. 
% Similarly, approaches like \textit{MuRAG} \citep{chen2022murag} and \textit{MultimodalQA} \citep{talmor2021multimodalqa} require joint reasoning over images and text to answer queries effectively.

% However, a critical bottleneck identified in recent surveys \citep{zhao2023retrieving} is the lack of a unified perception stage.
% Images and text are often processed in isolation, breaking the semantic flow of the document.
% Furthermore, standard Optical Character Recognition (OCR) often loses the structural integrity of complex tables.
% MiRAGE resolves this by implementing a multimodal ingestion \& segmentation phase. 
% It utilizes a vision agent capable of generating image descriptions and extracting table structures into an enriched markdown, combined with a segmentation agent performing semantic chunking.
% Our framework preserves the semantic proximity between textual and visual elements, creating a shared vector space that mirrors the source document's layout.

\subsection{Multi-hop Reasoning}

Effective RAG systems shall go beyond simple information retrieval to synthesize information across scattered but semantically related content. 
% different document sections cross-referencing each other and semantic references.
Research into \textit{MultiHop-RAG} \citep{tang2024multihop} and \textit{HotpotQA} \citep{yang2018hotpotqa} has demonstrated that systems performing well on single-hop queries often fail significantly when required to connect multiple pieces of evidence. 
While datasets like \textit{MuSiQue} \citep{trivedi2022musique} and \textit{Graphhopper} \citep{koner2021graphhopper} introduce compositional queries, they largely rely on open-domain sources (e.g., Wikipedia) or extractive answers.
% \textit{MuSiQue} \citep{trivedi2022musique} and \textit{Graphhopper} \citep{koner2021graphhopper} further advanced this direction by introducing compositional queries that require sequential reasoning steps. % or navigation through scene graphs.
% A major limitation of existing multi-hop datasets is their reliance on open-domain sources like Wikipedia (e.g., \textit{Wiki-MultihopQA} \citep{ho-etal-2020-constructing}) or their extractive nature, where answers are direct spans of text rather than synthesized responses.
Crucially, they lack the specific expert persona required for domain-specific tasks, such as engineering or finance \citep{schnitzler2024morehopqa}.
MiRAGE automates the creation of complex reasoning chains through a context optimization loop. 
The context agent iteratively expands search queries to gather scattered evidence, allowing the QA generator agent to create complex, multi-hop Q\&A pairs that mimic expert analysis rather than simple keyword matching.

% Systems optimized for single-hop retrieval often fail to connect scattered evidence \citep{tang2024multihop, yang2018hotpotqa}. 
% While datasets like \textit{MuSiQue} \citep{trivedi2022musique} and \textit{Graphhopper} \citep{koner2021graphhopper} introduce compositional queries, they largely rely on open-domain sources (e.g., Wikipedia) or extractive answers.
% Crucially, they lack the specific expert persona required for domains like finance or engineering \citep{schnitzler2024morehopqa}. 
% Unlike these extractive approaches, MiRAGE utilizes a context optimization loop where a context agent iteratively expands search queries. 
% This allows the generation of multi-hop QA pairs that require synthesizing disjoint evidence, mimicking expert cognitive workflows rather than simple span extraction.

\begin{figure*}[t]
    \centering
    % \fbox{%
        \includegraphics[
            width=0.9\linewidth,
            trim=0in 0.1in 0in 0in, % l b r t  (IMPORTANT order)
            clip
        ]{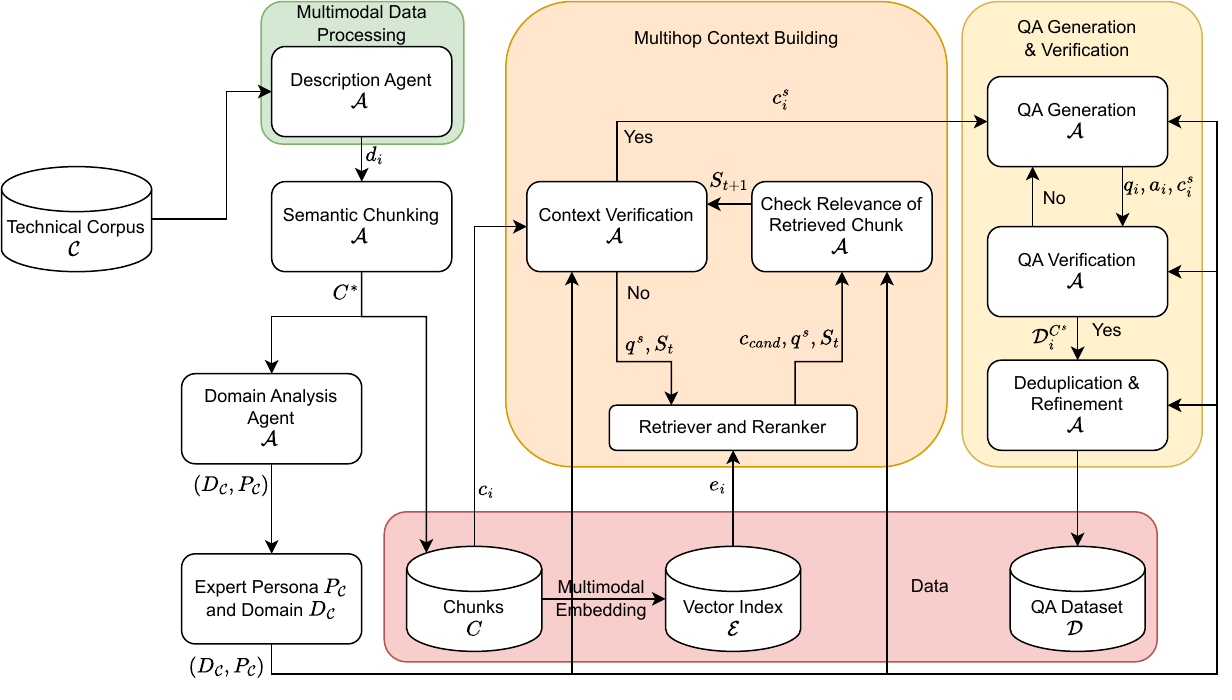}
    % }
    \vspace{-0.1in}

    \caption{The multiagent framework of MiRAGE to evaluate RAG systems}
    \vspace{-0.2in}
    \label{fig:agentic_architecture}
\end{figure*}

\subsection{Agentic Frameworks for Dataset Generation}

The complexity of generating verified, multimodal, multi-hop questions has led to the emergence of agentic frameworks. 
\textit{SMMQG} \citep{wu2024synthetic} employs an interplay between a retriever, LLM, and a multimodal model to synthesize questions. 
Similarly, \textit{WeQA} \citep{meyur2025weqa} utilize Human-AI teaming to ensure domain relevance.
However, the linear pipeline of \textit{SMMQG} \citep{wu2024synthetic} lacks a robust feedback mechanism for deduplication, often resulting in semantic redundancy and inflated datasets. 
% for deduplication and quality control at scale.
% Linear pipelines often produce repetitive or semantically overlapping questions, inflating dataset size without adding value. 
MiRAGE advances this paradigm by organizing agents into a generation swarm.
By incorporating a selector agent to filter for difficulty and a curator agent that performs hierarchical deduplication and clustering, MiRAGE ensures the final output is not only accurate but diverse, minimizing redundancy while maximizing semantic coverage.

% The complexity of generating verified, multimodal questions has led to agentic frameworks like \textit{SMMQG} \citep{wu2024synthetic}, which employs an interplay between retrievers and multimodal models, and \textit{WeQA} \citep{meyur2025weqa}, which utilizes Human-AI teaming.
% However, \textit{SMMQG} utilizes a linear pipeline that lacks robust feedback mechanisms for deduplication, often resulting in semantic redundancy. 
% MiRAGE advances this paradigm by organizing agents into a generation swarm. 
% By incorporating a selector agent for difficulty filtering and a curator agent for hierarchical clustering, MiRAGE ensures the output is diverse and maximizes semantic coverage while minimizing redundancy.

\section{Methodology}
\label{sec:methodology}
We propose a multimodal multi-agentic framework designed to generate a question-answer dataset $(q_i, a_i) \in \mathcal{D}$ from a technical corpora $\mathcal{C}$.
$\mathcal{D}$ is a collection of validated pairs $(q_i, a_i)$. 
Mathematically, the dataset generation process is defined by
\begin{equation}
\mathcal{D} = \Phi \left( \left\{ (q_i, a_i) \sim P(q, a | V=1, \mathcal{C}, \Theta) \right\}_{i=1}^N \right)
\end{equation}
The joint probability of generating a valid pair from its context is
\begin{equation}
\begin{split}
    P(q, a, & V=1 | \mathcal{C}, \Theta) = \\
     \sum_{C^s \subseteq \mathcal{C}} \Big[ & P(V=1 | q, a, C^s, \theta_V) \\
     & \cdot P(q,a | C^s, \theta_{QA}) \\
     & \cdot P(C^s | \mathcal{C}, \theta_{S}) \Big]
\end{split}
\label{eq:generative_process}
\end{equation}
where $C^s$ is a semantic context (a subset of chunks) derived from $\mathcal{C}$.
The parameter set $\Theta = \{\theta_V, \theta_{QA}, \theta_{S}\}$ represents the configurations for the verification, generation, and semantic search agents, respectively.
The framework operates in five phases:
(1) multimodal data ingestion and semantic chunking,
(2) identification of expert persona and domain,
(3) semantic multihop context building, 
(4) agentic QA generation and verification, and lastly
(5) refinement and deduplication.
The architecture is model-agnostic, allowing for the interchange of underlying language models.
The framework is illustrated in Fig. \ref{fig:agentic_architecture}.

\subsection{Multimodal Data Ingestion and Semantic Chunking}
Raw technical documents contain complex layouts where textual information is inextricably linked with visual artifacts. To address this, we implement a hybrid parsing pipeline to structure $\mathcal{C}$.
We utilize a document layout analysis engine for structural segmentation and a description agent powered by a Vision Language Model (VLM) for visual interpretation.
The description agent generates a dense textual description $d_i$ for every visual element $v_i \in \mathcal{C}$.
\begin{equation}
    d_i = \mathcal{A}(v_i \mid \pi_{desc})
\end{equation}
where $\pi_{desc}$ is the prompt to \textit{describe} the technical details.

We employ a semantic chunking strategy to segment the source document into semantically coherent chunks.
The document text $T$ is processed via a sliding window of length $L$ with an overlap of $l$.
For each window $W_t$, the semantic chunking agent identifies the optimal partition $C^*$ from all the possible partitions of $W_t$,
%\begin{equation}
%    C^* = \underset{C \in \text{Partitions}(W_t)} {\arg\min} \left( \sum_{j=1}^{|C|-1} \left[ 1 - \text{Sim}(c_j, c_{j+1}) \right] + \lambda |C| \right)
%\label{eq:chunk_dissimilarity}
%\end{equation}
\begin{equation}
\begin{split}
    C^* = \operatorname*{arg\,min}_{C \in \text{Partitions}(W_t)} \bigg( & \sum_{j=1}^{|C|-1} [1 - \text{Sim}(c_j, c_{j+1})] \\
    & + \lambda |C| \bigg)
\end{split}
\label{eq:chunk_dissimilarity}
\end{equation}
where $c_j, c_{j+1}$ are adjacent chunks in candidate partition $C$, and $\text{Sim}(\cdot)$ is a semantic similarity function.
$\lambda$ is a regularization hyperparameter that prevents over-fragmentation.
% The resulting chunks form the base corpus $\mathcal{C}$.

\subsection{Identification of Expert Persona and Domain}
\label{sec:domain_identification}
We incorporate an agent to determine the core domain $D_{\mathcal{C}}$ and an expert persona $P_{\mathcal{C}}$ by performing a global analysis of the corpus.
These parameters condition the generation agents to reflect the domain knowledge and the style and complexity of a subject matter expert.
Each chunk $c_i \in \mathcal{C}$ is represented by its multimodal embedding $e_i$. %  \in \mathbb{R}^d
Let $\mathcal{E} = \{e_1, \dots, e_M\}$ be the collection of embeddings.
The agent employs a topic modeling pipeline to discover the latent thematic structure within the corpus using three steps.
First, $\mathcal{E}$ is projected into a lower-dimensional space using manifold learning.
Second, density-based clustering partitions the chunks into $K$ thematic clusters $\{\tau_1, \dots, \tau_K\}$ where each cluster correspond to a distinct topic.
Third, we utilize class-based TF-IDF to generate a representative keyword list $R_k$ for each cluster $\tau_k$, refined using Maximal Marginal Relevance (MMR) to promote keyword diversity.
Finally, the agent synthesizes the domain and persona from the dominant topic representations $\{R_k\}$,
\begin{equation}
(D_{\mathcal{C}}, P_{\mathcal{C}}) \sim \mathcal{A}(D, P \mid \{R_k\}_{k=1}^K, \pi_{DP})
\label{eq:domain_synthesis}
\end{equation}
where $\pi_{DP}$ is the prompt to extract the domain of the corpus and a relevant expert persona.
% These derived values update $\theta_{QA}$ for subsequent phases.

\begin{table*}[t]
\centering
\caption{Summary of selected corpora for MiRAGE evaluation.
% The selection prioritizes diversity in modality, semantic structure, and reasoning requirements.
}
\vspace{-0.1in}
\label{tab:corpora_stats}
\resizebox{1.2\columnwidth}{!}{%
\begin{tabular}{l l l l l l}
\toprule
\textbf{Dataset Name} & \textbf{Domain}  & \textbf{\# Pages} & \textbf{\# Images} & \textbf{\# Tables} & \textbf{\# Tokens} \\
\midrule
S\&P Global Annual Reports & Finance  & 1302 & 1,120 & 2,800 & 0.9M \\
UNECE GTRs & Regulation  & 7594 & 150 & 3,450 & 3.8M \\
Quantitative Biology & Science & 8336 & 9,400 & 850 & 5.2M \\
NYTimes Opinions & Journalism  & $>$3000 & 3,050 & 25 & 2.1M \\
\bottomrule
\vspace{-0.4in}
\end{tabular}%
}
\end{table*}

\subsection{Semantic Multihop Context Building}
\label{sec:multihop_context}
% This phase realizes the probability term $P(C^s | \mathcal{C}, \theta_{S})$ in Equation \ref{eq:generative_process}.
We implement a recursive, agentic retrieval process that expands an initial seed chunk $c_{seed}$ into a complete semantic context $C^s$.
Let $S_t$ denote the set of chunks constituting the context at iteration $t$, where $S_0 = \{c_{seed}\}$.
At step $t$, a multimodal agent analyzes $S_t$ to detect missing information.
We model this as a conditional generation task:
\begin{equation}
    (z_t, Q^{search}_t) = \mathcal{A}(S_t \mid \pi_{comp})
\end{equation}
where $\pi_{comp}$ is the completeness prompt.
The output consists of a boolean status $z_t$ (complete/incomplete) and, if incomplete, a set of retrieval queries $Q^{search}_t = \{q^{s}_1, \dots, q^{s}_k\}$.
The process terminates if $z_t=1$ or $t \geq \delta_{max}$, yielding $C^s = S_t$.
% For each query $q^{s} \in Q^{search}_t$, we retrieve top-$N$ candidates from $\mathcal{E}$ and re-rank them.
For each generated query $q^{s} \in Q^{search}_t$, we employ a hybrid retrieval strategy. 
We first retrieve the top-$N$ candidates from $\mathcal{E}$ followed by reranking by a multimodal reranker.

To prevent context drift, we verify if a candidate chunk $c_{cand}$ retrieved by the query $q^{s}$ specifically offers the information missing in $S_t$ or is related to $S_t$.
The context expands by
\begin{equation}
    S_{t+1} = S_t \cup \left\{ c_{cand} \mid \mathcal{A}(S_t, q^{s}, c_{cand} \mid \pi_{add}) \right\}
\end{equation}
where $\pi_{add}$ is the prompt that assesses the suitability of $c_{cand}$.
This ensures the construction of $C^s$ is strictly monotonic regarding information utility.

\subsection{QA Generation and Verification}
\label{sec:qa_generation}
We generate candidate question-answer pairs $(q, a)$ from the context $C^s$ and verify them. 
This corresponds to the probability terms $P(q,a | C^s, \theta_{QA})$ and $P(V=1 | q, a, C^s, \theta_V)$ defined in Eq. \ref{eq:generative_process}.
To ensure technical depth, we condition the QA generation on the context $C^s$, the domain $D_{\mathcal{C}}$, and the persona $P_{\mathcal{C}}$ (encapsulated in $\theta_{QA}$).
The set of candidates $\mathcal{D}_{cand}$ is generated as:
\begin{equation}
    \mathcal{D}_{cand} = \left\{ (q_i, a_i) \right\}_{i=1}^{M} \sim \mathcal{A}(C_i^s, P_{\mathcal{C}}, D_{\mathcal{C}} \mid \pi_{QA})
\end{equation}
where $\pi_{QA}$ is the QA generation prompt. 

To ensure that the agent is relying only on the provided context $C^s$, a verifier agent assesses each candidate $(q, a) \in \mathcal{D}_{cand}$ against $C^s$.
The verification function evaluates:
(1) Correctness: $a$ is factually supported by $C^s$.
(2) Necessity: $q$ requires information within $C^s$ to get $a$.
The validated dataset contribution from context $C^s$ is:
\begin{equation}
    \mathcal{D}_i^{C^s} = \left\{ (q, a) \in \mathcal{D}_{cand} \mid \mathcal{A}(q, a, C_i^s \mid \pi_{ver}) \right\}
\end{equation}
where $\pi_{ver}$ is the verification prompt.

\subsection{Refinement and Deduplication}
\label{sec:refinement}
The final stage aggregates the $\mathcal{D}_i^{C^s}$ into the final cohesive corpus $\mathcal{D}$ via hierarchical clustering and deduplication.
We employ a two-stage clustering process.
First, we apply community detection on the question embeddings to form a set of high-level clusters $\mathcal{K}_{Q}$ that share semantic themes.
Second, within each cluster $k \in \mathcal{K}_{Q}$, we sub-cluster the answers to identify redundancy. The union of these sub-clusters constitutes the fine-grained answer cluster set $\mathcal{K}_{A}$.

The similarity metric between pairs $u_i=(q_i, a_i)$ and $u_j=(q_j, a_j)$ is a weighted sum of semantic similarity and source lineage overlap:
\begin{equation}
    Sim(u_i, u_j) = \alpha \cos(e_{a_i}, e_{a_j}) + (1-\alpha) \cdot J(C^s_i, C^s_j)
\end{equation}
where $e_a$ is the answer embedding, and $J(\cdot)$ is the Jaccard similarity of the source contexts $C^s_i$ and $C^s_j$ associated with $u_i$ and $u_j$, respectively. $\alpha$ is a weighting factor.

Lastly, we employ a stratified policy to refine the QA pairs within each answer sub-cluster into representative units.
We define a refinement function $\Psi$ that operates on the QA pairs in each cluster $K \in \mathcal{K}_{A}$.
If the internal similarity of the cluster exceeds the threshold $\tau_t$, they are merged by the refinement agent; otherwise, the original units are retained. 
Mathematically,
\begin{equation}
    \Psi(K) = 
    \begin{cases} 
        \mathcal{A}(K \mid \pi_{ref}) & \text{if } \underset{u_i, u_j \in K}{\min} \, Sim(u_i, u_j) > \tau_t \\
        K & \text{otherwise}
    \end{cases}
\label{eq:stratified_policy}
\end{equation}
where $\pi_{ref}$ is the prompt instructing the agent to synthesize a unified refined set of QA pairs that encompasses the unique details from all QA pairs in $K$.
The final optimized dataset $\mathcal{D}$ is constructed by the union of all the refined QA pairs:
\begin{equation}
    \mathcal{D} = \bigcup_{K \in \mathcal{K}_{A}} \Psi(K)
\end{equation}

\section{Experiments}

\begin{figure*}[t]
    \centering
    % \fbox{%
        \includegraphics[
            width=\linewidth,
            trim=0.3in 1.5in 1.2in 1.1in, % l b r t  (IMPORTANT order)
            clip
        ]{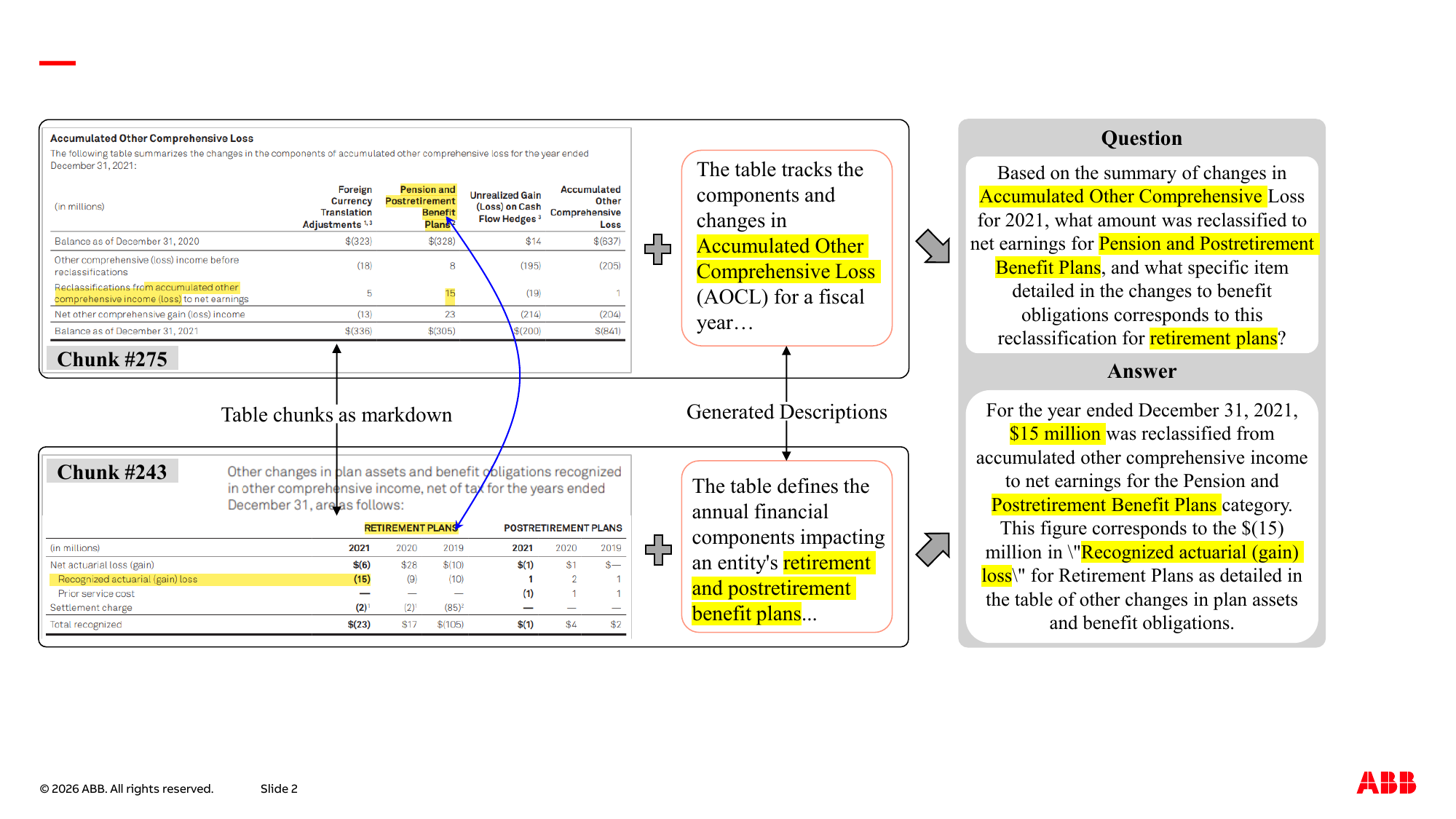}
    % }
    
    \caption{A sample question-answer pair generated from two related chunks with relevant keywords highlighted}
    \label{fig:exampleqa}
    \vspace{-0.2in}

\end{figure*}

\subsection{Corpora Selection}
To validate the efficacy of \textbf{MiRAGE} across the full spectrum of enterprise and information retrieval challenges, we curated a diverse suite of corpora spanning four distinct domains: finance, regulations, science, and journalism.
The \textit{S\&P Global Annual Reports} \cite{spglobal_annual_reports} represent the financial domain, where strategic narratives are deeply intertwined with dense tabular data and financial visualizations.
The \textit{UNECE Global Technical Regulations} \cite{unece_gtrs} are characterized by rigid hierarchical structures, precise definitions, and conditional logic.
% that dictates vehicle safety standards. It challenges the framework to generate queries requiring exact retrieval and logical verification of compliance criteria (e.g., specific load test forces or door retention component requirements), testing the verifier agent's ability to enforce strict adherence to regulatory clauses.
% It serves as a stress test for the multimodal semantic chunking agent, requiring the synthesis of information across modalities---such as correlating CEO outlooks with GAAP financial tables---to generate valid multi-hop queries that reflect real-world financial analysis.
We selected scientific publications from the quantitative biology domain submitted to the Arxiv\footnote{\url{https://arxiv.org/list/q-bio/2025-01?skip=0&show=2000}} in the month of January of 2025.
This corpus introduces extreme lexical specificity regarding protein structures and complex visual artifacts like 3D molecular renderings.
% It validates the framework's capacity to adopt an expert persona capable of formulating dense technical inquiries about specific methodologies (e.g., RA-EGN architecture) and interpreting domain-specific figures, moving beyond surface-level keyword matching to deep scientific reasoning.
The last dataset consists of the Opinions from The New York Times collected from the Visual News dataset \cite{liu2021visual}.
The datasets are listed in Table \ref{tab:corpora_stats}.
% Its generic nature requires grounding abstract textual captions in specific visual evidence within "in-the-wild" imagery. This selection evaluates the framework's versatility in handling associative image-text relationships and tests the vision agent's ability to distinguish between visual hallucination and grounded description across a broad spectrum of real-world events and entities.
These domains provide a comprehensive validation ground, moving from highly structured text-heavy documents to unstructured visually complex media. 

\subsection{Model Selection}
The MiRAGE framework is architected to be model-agnostic, supporting a modular interchange of LLMs and VLMs.
Our architecture primarily leverages state-of-the-art proprietary models to ensure maximum reasoning fidelity.
% \textbf{Generative Agents (LLMs \& VLMs):}
We utilize Gemini-2.5-Flash and GPT-5-Mini as the core engines for the reasoning agents.
Their high-context windows, advanced vision encoding, and superior instruction-following capabilities are critical for the multi-hop QA generation and verification tasks.
% \textbf{Embedding Models:}
The multimodal chunks are embedded with the Nomic model \cite{nussbaum2024nomic}
for semantic retrieval based on textual queries\footnote{The token limits of the embedding and reranker models based on the CLIP and SigLIP architectures render them ineffective for our multimodal tasks.}.
% Retrieval precision is refined using a two-stage reranking process.
The retrieval precision during the multihop context building is refined with multimodal rerankers.
We employ the LLM-as-a-reranker \cite{abdallah2025good} paradigm utilizing the proprietary models to rank the retrieved chunks.
This ensures that the final context window contains only the most semantically pertinent chunks.

\subsection{Metrics}
\label{sec:metrics}

To rigorously evaluate the quality of the generated dataset $\mathcal{D}$, we employ a comprehensive suite of metrics that assesses retrieval accuracy, reasoning complexity, multimodal integration, and domain coverage. 
We utilize both standard automated metrics \cite{es2024ragas} and novel agentic evaluation protocols based on the LLM-as-a-Judge paradigm \citep{zheng2023judging}.

For each QA pair $(q, a)$ and retrieved context $C^s$, we measure the faithfulness and relevance.
% , and the precision and recall of the retrieved context.
To validate that $\mathcal{D}$ necessitates multi-hop context building rather than simple extraction, we implement a reasoning trace evaluator $\mathcal{T}(q, C^s)$.
It consists of hop count ($H$), number of distinct retrieval or reasoning steps required to derive $a$ from $(q, C^s)$.
% The reasoning trace is supplemented with a reasoning score $S_{reason}\in [0,1]$ representing the complexity of the synthesis. 
% This discriminates between queries that aggregate independent facts (lower score) and those requiring logical deduction (e.g., comparison, causality) across disjoint chunks (higher score).
We employ a VLM agent as a verifier to ensure the generated answer $a$ is visually grounded.
The agent verifies that specific visual features referenced in $a$ (e.g., trend lines, molecular structures) are present in the image set $V \subset C^s$.
% \paragraph{Domain Coverage:}
We quantify the alignment between the latent topics of the source corpus and the generated QA dataset using the Jensen-Shannon (JS) divergence.
Let $P_{\mathcal{C}}$ and $P_{\mathcal{D}}$ be the discrete probability distributions over latent topic clusters for the corpus and dataset, respectively. 
The coverage metric is defined as:
\begin{equation}
    D_{JS}(P_{\mathcal{C}} || P_{\mathcal{D}}) = \frac{1}{2} D_{KL}(P_{\mathcal{C}} || M) + \frac{1}{2} D_{KL}(P_{\mathcal{D}} || M)
\end{equation}
where $M = \frac{1}{2}(P_{\mathcal{C}} + P_{\mathcal{D}})$. 
Lower divergence values imply that the synthetic dataset faithfully reproduces the thematic distribution of the original domain.

\section{Results}

We present the empirical evaluation of the QA datasets generated by MiRAGE.
We create 1000 QA pairs from each corpus.
% The prompts \textcolor{blue}{excluding the domain specific few shot examples} are included in Appendix \ref{sec:appendix_prompts}.
The prompts excluding the domain specific few shot examples are included in Appendix \ref{sec:appendix_prompts}.
A sample QA pair is illustrated in Figure \ref{fig:exampleqa}.
Our analysis focuses on the complexity of reasoning, the faithfulness of the generated answers, and the semantic alignment with the source domains.

\begin{table*}[t]
\centering
\caption{Performance comparison of MiRAGE across four domains.}
\vspace{-0.15in}
\label{tab:main_results}
\resizebox{0.6\textwidth}{!}{%
\begin{tabular}{l l c c c c}
\toprule
\textbf{Dataset} & \textbf{Model} & \textbf{Faith.} $\uparrow$ & \textbf{Rel.} $\uparrow$ & \textbf{Avg Hops ($H$)} $\uparrow$ & \textbf{Vis. Gr.} $\uparrow$ \\
\midrule
\multirow{2}{*}{\begin{tabular}{@{}l@{}}\textbf{S\&P Global} \\ (Finance)\end{tabular}}
% & Linear Baseline & 0.55 & 0.78 & 1.10 & 0.38 \\
% & Qwen3-VL:32b & 0.85 & 0.89 & 2.30 & 0.19 \\
& Gemini 2.5 Flash & \textbf{0.96} & \textbf{0.86} & \textbf{2.84} & 0.21 \\
& GPT 5 Mini & 0.91 & 0.81 & 2.42 & \textbf{0.28} \\
\midrule
\multirow{2}{*}{\begin{tabular}{@{}l@{}}\textbf{UNECE GTRs} \\ (Regulation)\end{tabular}}
% & Linear Baseline & 0.68 & 0.82 & 1.05 & 0.45 \\
% & Qwen3-VL:32b & 0.89 & 0.91 & 2.15 & 0.18 \\
& Gemini 2.5 Flash & 0.94 & \textbf{0.90} & 2.45 & 0.38 \\
& GPT 5 Mini & \textbf{0.96} & 0.82 & \textbf{2.60} & \textbf{0.45} \\
\midrule
\multirow{2}{*}{\begin{tabular}{@{}l@{}}\textbf{Q-Bio Arxiv} \\ (Science)\end{tabular}}
% & Linear Baseline & 0.61 & 0.75 & 1.02 & 0.41 \\
% & Qwen3-VL:32b & 0.82 & 0.85 & 1.95 & 0.21 \\
& Gemini 2.5 Flash & \textbf{0.83} & 0.92 & 2.35 & \textbf{0.42} \\
& GPT 5 Mini & 0.81 & \textbf{0.94} & \textbf{2.55} & 0.42 \\
\midrule
\multirow{2}{*}{\begin{tabular}{@{}l@{}}\textbf{NYTimes} \\ (Journalism)\end{tabular}}
% & Linear Baseline & 0.72 & 0.88 & 1.08 & 0.29 \\
% & Qwen3-VL:32b & 0.88 & 0.91 & 1.85 & 0.17 \\
& Gemini 2.5 Flash & 0.91 & 0.93 & 1.10 & \textbf{0.32} \\
& GPT 5 Mini & \textbf{0.93} & \textbf{0.95} & \textbf{1.25} & 0.24 \\
\bottomrule
\end{tabular}%
}
\vspace{-0.15in}

\end{table*}

\subsection{Overall Performance}
Table \ref{tab:main_results} summarizes the performance of the MiRAGE framework powered by two distinct state-of-the-art models: Gemini 2.5 Flash and GPT 5 Mini.
Our analysis highlights the framework's ability to generate complex, domain-aligned question-answer pairs while revealing nuanced differences in model capabilities across various tasks. 
The JS divergence from Table \ref{tab:ablation_combined} reflects that the QA dataset $\mathcal{D}$ generated by MiRAGE covers the topics in $\mathcal{C}$ effectively.

\paragraph{Reasoning Complexity:}
A key achievement of MiRAGE is its consistent generation of multi-hop questions across technical domains. 
For the Finance, Regulation, and Science corpora, the average hop count ($H$) consistently exceeds 2.3, peaking at 2.84 with Gemini 2.5 Flash on the S\&P Global dataset. 
This demonstrates the efficacy of the semantic multihop context-building phase, which successfully forces models to synthesize information from disjoint sources. 
GPT 5 Mini shows a slight advantage in generating more complex reasoning chains in the highly structured regulatory and scientific domains. 
In contrast, the lower hop count for the NYTimes corpus (avg. $H \approx 1.2$) reflects the open and less connected nature of journalistic content. 
The NYTimes corpus did not contain the chunks relevant to the queries to make the context complete.

\paragraph{Faithfulness and Relevance:}
MiRAGE demonstrates exceptional performance in maintaining factual grounding and contextual relevance, with faithfulness scores consistently above 0.91 for three of the four domains and relevance scores exceeding 0.81 across all experiments. 
This validates the effectiveness of the adversarial verifier agent, which successfully filters out hallucinations and ensures that generated answers are strictly supported by the retrieved context. 
Both Gemini 2.5 Flash and GPT 5 Mini perform competitively with marginal differences.
% indicating that the framework's architectural safeguards are robust and not overly dependent on a single model's characteristics.

\paragraph{Visual Grounding:}
While excelling in textual reasoning, the evaluation exposes visual grounding as a persistent challenge for current VLMs. 
The visual grounding scores remain moderate across all domains, with a maximum of 0.45 achieved by GPT 5 Mini on the UNECE GTRs dataset.
This suggests that while models can describe images, generating complex questions that require precise reasoning about specific visual elements (E.g., correlating specific data points in a financial chart) remains a frontier.
Out of the 1093 QA pairs generated for the finance domain, only 84 QA pairs are multimodal. It indicates that the VLMs prefer textual content to generate the QA pairs.
From a closer look at the multimodal QA pairs, we hypothesize that the generated descriptions made the visual elements partially redundant leading to a consistently lower score visual grounding score across all evaluations.  

\subsection{Ablation Study}

\begin{table*}[t]
\centering
\caption{Ablation study on the S\&P Global Annual Reports (Finance) dataset}
\label{tab:ablation_combined}
\vspace{-0.15in}

\resizebox{0.8\textwidth}{!}{%
\begin{tabular}{l l c c c c c c}
\toprule
\textbf{Category} & \textbf{Configuration} & \textbf{Faith.} $\uparrow$ & \textbf{Rel.} $\uparrow$ & \textbf{Diff} $\uparrow$ & \textbf{Avg Hops} $\uparrow$ & \textbf{Vis. Gr.} $\uparrow$ & \textbf{JSD} $\downarrow$ \\
\midrule
% \rowcolor{gray!10} 
\textbf{Baseline} & \textbf{MiRAGE} & \textbf{0.97} & \textbf{0.95} & \textbf{0.85} & 1.92 & 0.41 & 0.08 \\
\midrule
\multirow{3}{*}{\textit{Agentic Architecture}} 
& (-) Multihop Context & 0.93 & 0.82 & 0.61 & - & 0.21 & \underline{4.35} \\
& (-) QA Verifier Agent & 0.74 & \underline{0.76} & 0.62 & 1.67 & 0.32 & \textbf{0.06} \\
& (-) Domain \& Persona & 0.91 & 0.88 & \underline{0.52} & 1.81 & \underline{0.11} & 0.11 \\
\midrule
\multirow{3}{*}{\textit{Data Representation}} 
& Fixed Chunk Size (2048) & 0.84 & 0.79 & 0.70 & \textbf{2.01} & 0.26 & 0.15 \\
& Image Only (No Description) & \underline{0.71} & 0.79 & 0.73 & \underline{1.34} & \textbf{0.62} & 1.73 \\
& Description Only (No Image) & 0.93 & 0.89 & 0.78 & 1.72 & - & 2.12 \\
\bottomrule
\end{tabular}%
}
\vspace{-0.15in}
\end{table*}

We conducted a systematic component-wise ablation using a subset (2 out of 14 documents) of the \textit{S\&P Global} annual reports. 
This corpus was selected for its high density of visual artifacts and complex tabular structures.
% , which provide a challenging testbed for multimodal integration. 
We isolate the impact of the agentic architecture and the data representation strategies. The results are summarized in Table \ref{tab:ablation_combined}. 

\subsubsection{Impact of Agentic Architecture}
% We evaluated the framework by individually removing the multihop context loop, the verifier agent, and the domain/persona injection. 

\paragraph{Multihop Context:}
Removing the recursive retrieval step forces the model to generate questions based solely on the initial seed chunk. 
This results in a collapse of the difficulty score\footnote{The domain specific expert agent rates the generated QA pair on a discrete scale of $[0,10]$, later normalized to $[0,1]$ for easier interpretation.} ($0.85 \rightarrow 0.61$). 
% Furthermore, the system fails to generate valid multi-hop queries (Avg Hops is undefined/single-step), confirming that without iterative expansion, 
Without the multihop context, the generation reverts to simple extractive QA yielding a degradation in domain alignment ($JSD \approx 4.35$) as well.
% , suggesting that single-hop contexts fail to capture the thematic distribution of complex financial documents. 

\paragraph{Verifier Agent:}
The removal of the adversarial verifier causes the most significant drop in faithfulness ($0.97 \rightarrow 0.74$).
Qualitative analysis reveals that the generation agent frequently hallucinates relationships between unconnected data points to satisfy the prompt's complexity requirements.
The drop in Relevance ($0.95 \rightarrow 0.76$) further indicates that unverified questions often drift from the core semantic content of the context. 

\paragraph{Domain/Persona Injection:}
Ablating the domain analysis prevents the model from adopting the specific expert persona (e.g., `Financial Reporting Analyst' for the S\&P Global annual reports) and relevant domain (`Corporate Financial Reporting and Analysis').
The Difficulty of the generated questions drops significantly ($0.85 \rightarrow 0.52$).
This indicates that without domain/persona conditioning, the model defaults to generic QAs rather than the deep, deductive reasoning characteristic of domain experts.

\subsubsection{Impact of Data Representation}
% We assessed the structural decisions regarding text segmentation and multimodal integration strategies. 

\paragraph{Chunking Strategy:}
We replaced the agentic semantic chunking (Eq. \ref{eq:chunk_dissimilarity}) with a standard fixed-window approach (2048 tokens). 
As shown in Table \ref{tab:ablation_combined}, fixed chunk size marginally increased the average hops ($1.92 \rightarrow 2.01$).
The marginal change in the average hops makes it inconclusive to highlight the importance of semantic chunking to avoid fragmentation of tables and semantically continuous text blocks.
However, the drop in the faithfulness and relevance allude to the importance of semantic chunking. 

\paragraph{Multimodal Configurations:}
We compared the hybrid MiRAGE approach against \textit{image Only} (no descriptions) and \textit{description Only} (no raw images) configurations.
The \textit{Image Only} setting results in higher visual grounding ($0.62$) indicating the increased importance of images.
However, it is accompanied by the lowest faithfulness ($0.71$) and lower relevance indicating the struggle of VLMs to bridge the semantic gap between the images and the text.
Conversely, the \textit{description Only} setting performs comparably with the full MiRAGE indicating that LLMs can power MiRAGE if descriptions of the visual artifacts are available.

\section{Conclusion}
We present \textbf{MiRAGE}, a comprehensive multi-agent framework that automates the generation of high-fidelity, multimodal, multi-hop evaluation datasets, addressing the critical need of domain-specific benchmarks for high-stakes RAG applications. 
By orchestrating a swarm of specialized agents through a recursive \textit{context optimization loop}, \textit{adversarial verification}, and \textit{domain-expert recognition}, MiRAGE successfully transcends the limitations of linear synthetic pipelines, producing datasets that exhibit deep deductive reasoning and strict factual adherence across diverse corpora.
Our empirical evaluation of the finance, science, regulations and journalism domains confirm the efficacy of MiRAGE in preserving semantic dependencies within complex technical documents.
Our ablation studies reveal that current state-of-the-art VLMs still rely significantly on dense textual descriptions to bridge the visual reasoning gap.
Ultimately, MiRAGE establishes a robust standard for automated dataset generation, empowering organizations to rigorously stress-test RAG systems against the latent complexity and specific thematic distributions of their proprietary data.

\vspace{-0.1in}

% The advancement of Multimodal RAG requires evaluation benchmarks that match the structural complexity of real-world information landscapes. 
% We present \textbf{MiRAGE}, a framework that evolves synthetic dataset generation by orchestrating a swarm of specialized agents to create verified, multi-hop, and multimodal QA pairs. 
% By replacing standard single-pass generation with a recursive context expansion and adversarial verification loop, MiRAGE ensures that generated queries require deep semantic synthesis across disjoint modalities rather than simple surface-level matching. 
% Our experiments across financial, scientific, journalism and regulatory domains demonstrate that this agentic approach yields benchmarks with superior reasoning complexity and visual grounding. 
% Ultimately, MiRAGE provides the essential infrastructure to rigorously validate and improve the reliability of information retrieval systems intended for high-stakes enterprise deployment.

\section*{Limitations and Future Work}
While MiRAGE advances the quality of synthetic benchmarks, it is not without limitations. 
The multiagentics architecture is computationally intensive.
The multihop context building and QA verification loops result in higher token costs and latency.
% Second, the quality of visual grounding remains bounded by the capabilities of the underlying VLM; while effective for technical diagrams, the description agents may still struggle with highly abstract or metaphorical imagery found in unconstrained open domains. 
Future work will focus on optimizing the agentic workflow for token efficiency.
Exploring the performance of open-source models,
% such as Qwen3-VL:32B \cite{bai2025qwen3vltechnicalreport},
would help democratize the framework.
% Lastly, a key foundation to multihop QA generation
% and extending the framework to support temporal reasoning over video and audio modalities.

% % Bibliography entries for the entire Anthology, followed by custom entries
% %\bibliography{anthology,custom}
% % Custom bibliography entries only
\bibliography{references}

@inproceedings{es2024ragas,
  title={Ragas: Automated evaluation of retrieval augmented generation},
  author={Es, Shahul and James, Jithin and Anke, Luis Espinosa and Schockaert, Steven},
  booktitle={Proceedings of the 18th Conference of the European Chapter of the Association for Computational Linguistics: System Demonstrations},
  pages={150--158},
  year={2024}
}

@article{kwiatkowski2019natural,
  title={Natural questions: a benchmark for question answering research},
  author={Kwiatkowski, Tom and Palomaki, Jennimaria and Redfield, Olivia and Collins, Michael and Parikh, Ankur and Alberti, Chris and Epstein, Danielle and Polosukhin, Illia and Devlin, Jacob and Lee, Kenton and others},
  journal={Transactions of the Association for Computational Linguistics},
  volume={7},
  pages={453--466},
  year={2019},
  publisher={MIT Press}
}

@inproceedings{abdallah2025good,
    title = "How Good are {LLM}-based Rerankers? An Empirical Analysis of State-of-the-Art Reranking Models",
    author = "Abdallah, Abdelrahman  and
      Piryani, Bhawna  and
      Mozafari, Jamshid  and
      Ali, Mohammed  and
      Jatowt, Adam",
    editor = "Christodoulopoulos, Christos  and
      Chakraborty, Tanmoy  and
      Rose, Carolyn  and
      Peng, Violet",
    booktitle = "Findings of the Association for Computational Linguistics: EMNLP 2025",
    month = nov,
    year = "2025",
    address = "Suzhou, China",
    publisher = "Association for Computational Linguistics",
    url = "https://aclanthology.org/2025.findings-emnlp.305/",
    doi = "10.18653/v1/2025.findings-emnlp.305",
    pages = "5693--5709",
    ISBN = "979-8-89176-335-7",
}

@article{nguyen2016ms,
  title={Ms marco: A human-generated machine reading comprehension dataset},
  author={Nguyen, Tri and Rosenberg, Mir and Song, Xia and Gao, Jianfeng and Tiwary, Saurabh and Majumder, Rangan and Deng, Li},
  year={2016}
}

@inproceedings{filice2025generating,
  title={Generating Q\&A benchmarks for RAG evaluation in enterprise settings},
  author={Filice, Simone and Horowitz, Guy and Carmel, David and Karnin, Zohar and Lewin-Eytan, Liane and Maarek, Yoelle},
  booktitle={Proceedings of the 63rd Annual Meeting of the Association for Computational Linguistics (Volume 6: Industry Track)},
  pages={469--484},
  year={2025}
}

@inproceedings{zheng2025revolutionizing,
  title={Revolutionizing Database Q\&A with Large Language Models: Comprehensive Benchmark and Evaluation},
  author={Zheng, Yihang and Li, Bo and Lin, Zhenghao and Luo, Yi and Zhou, Xuanhe and Lin, Chen and Li, Guoliang and Su, Jinsong},
  booktitle={Proceedings of the 31st ACM SIGKDD Conference on Knowledge Discovery and Data Mining V. 2},
  pages={5960--5971},
  year={2025}
}

@inproceedings{sivasothy2025ragprobe,
  title={RAGProbe: Breaking RAG Pipelines with Evaluation Scenarios},
  author={Sivasothy, Shangeetha and Barnett, Scott and Kurniawan, Stefanus and Rasool, Zafaryab and Vasa, Rajesh},
  booktitle={2025 IEEE/ACM 4th International Conference on AI Engineering--Software Engineering for AI (CAIN)},
  pages={60--71},
  year={2025},
  organization={IEEE}
}

@article{kim2025visual,
  title={Visual question answering: A survey of methods, datasets, evaluation, and challenges},
  author={Kim, Byeong Su and Kim, Jieun and Lee, Deokwoo and Jang, Beakcheol},
  journal={ACM Computing Surveys},
  volume={57},
  number={10},
  pages={1--35},
  year={2025},
  publisher={ACM New York, NY}
}

@article{mei2025survey,
  title={A survey of multimodal retrieval-augmented generation},
  author={Mei, Lang and Mo, Siyu and Yang, Zhihan and Chen, Chong},
  journal={arXiv preprint arXiv:2504.08748},
  year={2025}
}

@article{pramanick2024spiqa,
  title={Spiqa: A dataset for multimodal question answering on scientific papers},
  author={Pramanick, Shraman and Chellappa, Rama and Venugopalan, Subhashini},
  journal={Advances in Neural Information Processing Systems},
  volume={37},
  pages={118807--118833},
  year={2024}
}

@article{foroutan2025wikimixqa,
  title={WikiMixQA: A Multimodal Benchmark for Question Answering over Tables and Charts},
  author={Foroutan, Negar and Romanou, Angelika and Ansaripour, Matin and Eisenschlos, Julian Martin and Aberer, Karl and Lebret, R{\'e}mi},
  journal={arXiv preprint arXiv:2506.15594},
  year={2025}
}

@inproceedings{chen2022murag,
  title={Murag: Multimodal retrieval-augmented generator for open question answering over images and text},
  author={Chen, Wenhu and Hu, Hexiang and Chen, Xi and Verga, Pat and Cohen, William},
  booktitle={Proceedings of the 2022 Conference on Empirical Methods in Natural Language Processing},
  pages={5558--5570},
  year={2022}
}

@article{zhao2023retrieving,
  title={Retrieving multimodal information for augmented generation: A survey},
  author={Zhao, Ruochen and Chen, Hailin and Wang, Weishi and Jiao, Fangkai and Do, Xuan Long and Qin, Chengwei and Ding, Bosheng and Guo, Xiaobao and Li, Minzhi and Li, Xingxuan and others},
  journal={arXiv preprint arXiv:2303.10868},
  year={2023}
}

@article{tang2024multihop,
  title={Multihop-rag: Benchmarking retrieval-augmented generation for multi-hop queries},
  author={Tang, Yixuan and Yang, Yi},
  journal={arXiv preprint arXiv:2401.15391},
  year={2024}
}

@inproceedings{yang2018hotpotqa,
  title={HotpotQA: A dataset for diverse, explainable multi-hop question answering},
  author={Yang, Zhilin and Qi, Peng and Zhang, Saizheng and Bengio, Yoshua and Cohen, William and Salakhutdinov, Ruslan and Manning, Christopher D},
  booktitle={Proceedings of the 2018 conference on empirical methods in natural language processing},
  pages={2369--2380},
  year={2018}
}

@article{trivedi2022musique,
  title={{MuSiQue}: Multihop Questions via Single-hop Question Composition},
  author={Trivedi, Harsh and Balasubramanian, Niranjan and Khot, Tushar and Sabharwal, Ashish},
  journal={Transactions of the Association for Computational Linguistics},
  volume={10},
  pages={539--554},
  year={2022},
  publisher={MIT Press One Broadway, 12th Floor, Cambridge, Massachusetts 02142, USA}
}

@inproceedings{koner2021graphhopper,
  title={Graphhopper: Multi-hop scene graph reasoning for visual question answering},
  author={Koner, Rajat and Li, Hang and Hildebrandt, Marcel and Das, Deepan and Tresp, Volker and G{\"u}nnemann, Stephan},
  booktitle={International Semantic Web Conference},
  pages={111--127},
  year={2021},
  organization={Springer}
}

@article{schnitzler2024morehopqa,
  title={Morehopqa: More than multi-hop reasoning},
  author={Schnitzler, Julian and Ho, Xanh and Huang, Jiahao and Boudin, Florian and Sugawara, Saku and Aizawa, Akiko},
  journal={arXiv preprint arXiv:2406.13397},
  year={2024}
}

@inproceedings{wu2024synthetic,
  title={Synthetic multimodal question generation},
  author={Wu, Ian and Jayanthi, Sravan and Viswanathan, Vijay and Rosenberg, Simon and Pakazad, Sina Khoshfetrat and Wu, Tongshuang and Neubig, Graham},
  booktitle={Findings of the Association for Computational Linguistics: EMNLP 2024},
  pages={12960--12993},
  year={2024}
}

@inproceedings{meyur2025weqa,
  title={Weqa: A benchmark for retrieval augmented generation in wind energy domain},
  author={Meyur, Rounak and Phan, Hung and Wagle, Sridevi and Strube, Jan and Halappanavar, Mahantesh and Horawalavithana, Sameera and Acharya, Anurag and Munikoti, Sai},
  booktitle={Proceedings of the Fourth Workshop on NLP for Positive Impact (NLP4PI)},
  pages={239--251},
  year={2025}
}

@article{lewis2020retrieval,
  title={Retrieval-augmented generation for knowledge-intensive nlp tasks},
  author={Lewis, Patrick and Perez, Ethan and Piktus, Aleksandra and Petroni, Fabio and Karpukhin, Vladimir and Goyal, Naman and K{\"u}ttler, Heinrich and Lewis, Mike and Yih, Wen-tau and Rockt{\"a}schel, Tim and others},
  journal={Advances in neural information processing systems},
  volume={33},
  pages={9459--9474},
  year={2020}
}

@article{nussbaum2024nomic,
  title={Nomic embed vision: Expanding the latent space},
  author={Nussbaum, Zach and Duderstadt, Brandon and Mulyar, Andriy},
  journal={arXiv preprint arXiv:2406.18587},
  year={2024}
}

@inproceedings{xiong2024benchmarking,
  title={Benchmarking retrieval-augmented generation for medicine},
  author={Xiong, Guangzhi and Jin, Qiao and Lu, Zhiyong and Zhang, Aidong},
  booktitle={Findings of the Association for Computational Linguistics ACL 2024},
  pages={6233--6251},
  year={2024}
}

@inproceedings{mallen2023not,
  title={When not to trust language models: Investigating effectiveness of parametric and non-parametric memories},
  author={Mallen, Alex and Asai, Akari and Zhong, Victor and Das, Rajarshi and Khashabi, Daniel and Hajishirzi, Hannaneh},
  booktitle={Proceedings of the 61st Annual Meeting of the Association for Computational Linguistics (Volume 1: Long Papers)},
  pages={9802--9822},
  year={2023}
}

@inproceedings{kandpal2023large,
  title={Large language models struggle to learn long-tail knowledge},
  author={Kandpal, Nikhil and Deng, Haikang and Roberts, Adam and Wallace, Eric and Raffel, Colin},
  booktitle={International conference on machine learning},
  pages={15696--15707},
  year={2023},
  organization={PMLR}
}

@article{zheng2023judging,
  title={Judging llm-as-a-judge with mt-bench and chatbot arena},
  author={Zheng, Lianmin and Chiang, Wei-Lin and Sheng, Ying and Zhuang, Siyuan and Wu, Zhanghao and Zhuang, Yonghao and Lin, Zi and Li, Zhuohan and Li, Dacheng and Xing, Eric and others},
  journal={Advances in neural information processing systems},
  volume={36},
  pages={46595--46623},
  year={2023}
}

@misc{unece_gtrs,
  author       = {{United Nations Economic Commission for Europe}},
  title        = {Global Technical Regulations ({GTRs})},
  howpublished = {\url{https://unece.org/transport/standards/transport/vehicle-regulations-wp29/global-technical-regulations-gtrs}},
  year         = {2023},
  note         = {Accessed: 2023-10-27}
}

@misc{spglobal_annual_reports,
  author       = {{S\&P Global}},
  title        = {Annual Reports},
  howpublished = {\url{https://investor.spglobal.com/sec-filings-reports/annual-reports/}},
  year         = {2023},
  note         = {Accessed: 2023-10-27}
}

@inproceedings{liu2021visual,
  title={Visual news: Benchmark and challenges in news image captioning},
  author={Liu, Fuxiao and Wang, Yinghan and Wang, Tianlu and Ordonez, Vicente},
  booktitle={Proceedings of the 2021 conference on empirical methods in natural language processing},
  pages={6761--6771},
  year={2021}
}

@inproceedings{park2025mirage,
  title={Mirage: A metric-intensive benchmark for retrieval-augmented generation evaluation},
  author={Park, Chanhee and Moon, Hyeonseok and Park, Chanjun and Lim, Heui-Seok},
  booktitle={Findings of the Association for Computational Linguistics: NAACL 2025},
  pages={2883--2900},
  year={2025}
}

\appendix

\section{Prompts}
\label{sec:appendix_prompts}

% ---------------------------------------------------------
\prompthead{PROMPTS_DESC["description"]}
\begin{lstlisting}
Provide a technical summary of the image/table for documentation.
Format: Single continuous paragraph, under 250 words. No bullets.
Content Requirements:
1. Identify the image type (table/figure) and its primary objective.
2. Technical Analysis: 
   - For Plots/Charts: Define axes/units, variables, and key trends/regions.
   - For Diagrams: Identify components, connection flows, and system boundaries.
   - Note: Describe visual attributes only if they encode data; ignore decorative elements and metadata.
3. Conclusion: Summarize critical insights and practical design implications.
\end{lstlisting}

% ---------------------------------------------------------
\prompthead{PROMPTS_CHUNK["semantic_chunking"]}
\begin{lstlisting}
You are a Semantic Chunking Engine. Segment markdown into coherent, verbatim chunks.
Processing Rules:
1. **Exclusions:** Ignore Table of Contents and Lists of Figures/Tables.
2. **Cohesion:** Merge orphan titles and short subsections into adjacent text. Ensure chunks are semantically self-contained.
3. **Status:** Mark as INCOMPLETE only if the final chunk ends abruptly (cut-off).
Chunk Classifications:
- **figure**: Image (`![...]`) with "Figure X" caption, description, and key. Set <artifact> to image path.
- **standalone image**: Image *without* "Figure X" caption. Set <artifact> to image path.
- **table**: "Table X" caption + markdown table + footnotes. Set <artifact> to 'None'.
- **table with images**: Table containing `![...]`. Set <artifact> to image path(s).
- **text**: Paragraphs, lists, or definitions. Set <artifact> to 'None'.
Output Format:
<chunk_id><|#|><chunk_type><|#|><content><|#|><artifact><|#|><status><|#|><chunk_end>
Field Definitions:
- chunk_id: Sequential integer starting at 1.
- chunk_type: text | table | table with images | figure | standalone image
- content: Exact unmodified markdown.
- artifact: Extracted image path(s) or `None'.
- status: COMPLETE | INCOMPLETE
\end{lstlisting}

% ---------------------------------------------------------
\prompthead{PROMPTS["domain_and_expert_from_topics"]}
\begin{lstlisting}
I have analyzed a technical document collection and extracted the following key topics:
{topic_list_str}
Based on these topics, please determine:
1. The specific technical or professional domain these topics belong to.
2. A specific expert role title for a professional in this domain.
Format your response exactly as follows (do not add any other text):
<|#|>START<|#|>
<|#|>Domain: <The Domain> 
<|#|>Expert Role: <The Expert Role>
<|#|>END<|#|>
\end{lstlisting}

% ---------------------------------------------------------
\prompthead{PROMPTS_CHUNK["completion_verification"]}
\begin{lstlisting}
You are a Chunk Completion Verification Agent. Evaluate if the provided text is semantically self-contained given that you are a(n) {expert_persona} working in the {domain} domain .
Criteria for INCOMPLETE status:
1. Missing Artifacts: References to Figures, Tables, or Sections not present in the chunk (e.g., "see Figure 1").
2. Undefined Context: Acronyms, technical terms, or variables used without definition or prior explanation.
3. Broken Continuity: Implicit references (e.g., "as mentioned above," "this method") or text describing a missing visual.
4. Rule: Do not assume expert inference. If a definition or artifact is missing, it is INCOMPLETE. Universal units are allowed.
Instructions:
- If COMPLETE: Confirm self-containment.
- If INCOMPLETE: Generate specific search queries to retrieve the missing definitions or artifacts.
Required Output Format:
Status: COMPLETE, Query: None, Explanation: <brief reasoning>
OR
Status: INCOMPLETE, Query: <specific_search_query_1> | <specific_search_query_2>, Explanation: <list missing refs/definitions>
\end{lstlisting}

% ---------------------------------------------------------
\prompthead{PROMPTS_CHUNK["chunk_addition_verification"]}
\begin{lstlisting}
You are a Chunk Addition Verification Agent ({expert_persona}, {domain}). Determine how a CANDIDATE CHUNK relates to an INCOMPLETE ORIGINAL CHUNK based on a specific SEARCH QUERY.
Classify as:
1. EXPLANATORY: Directly resolves the missing element. It provides the specific figure/table, defines the unknown term, or supplies the explicitly referenced prior context.
2. RELATED: Contextually relevant but does not solve the specific gap. Includes general theory, complementary data, or content useful for multi-hop QA (even if it references the same missing artifact).
3. UNRELATED: No semantic overlap or domain relevance.
Output Format:
Status: <EXPLANATORY | RELATED | UNRELATED>
Explanation: <Brief justification>
\end{lstlisting}

% ---------------------------------------------------------
\prompthead{PROMPTS["multi_hop_qa_generation"]}
\begin{lstlisting}
You are a(n) {expert_persona} in {domain_context}. Construct a high-quality Question-Answer pair by synthesizing information across the provided text chunks.
Content:
{content}
**Execution Protocol (Strict Order):**
1. **Chunk Count:** Identify how many distinct chunks are present.
2. **Keyword Extraction:** List critical technical keywords for *each* chunk.
3. **Relationship Mapping:** Identify "Bridge Keywords"---concepts that relate or intersect across the chunks.
4. **QA Synthesis:** Frame question-answer pairs that requires synthesizing these related keywords. The question must be unsolvable without combining info from multiple points.
5. **Decomposition:** Map specific parts of the Question and Answer back to their source chunks.
**Critical Constraints:**
- **No Hallucination:** Answer ONLY using provided content.
- **Self-Sufficiency:** The question must be standalone. NEVER use phrases like "the provided figure," "the text above," or "Section 2.1" without context. Explicitly name the object (e.g., "In Figure XX from document YY...").
- **Complexity:** The question must be multi-hop (requires connecting A to B).
**Output Format:**
<|#|>ANALYSIS<|#|>
Chunk Count: <Integer>
Keywords per Chunk: <Chunk 1: [A, B], Chunk 2: [C, D]>
Related Keywords: <[A] relates to [C] via...>
<|#|>QA_GENERATION<|#|>
Question: <Your specific, self-contained question>
Answer: <Concise technical answer>
Relevance: <0-10>
Difficulty: <0-10>
<|#|>DECOMPOSITION<|#|>
Question Source: <"Part of Q" -> derived from Chunk X>
Answer Source: <"Part of A" -> derived from Chunk Y>
<|#|>END<|#|>
\end{lstlisting}

% ---------------------------------------------------------
\prompthead{PROMPTS["question_answer_verification"]}
\begin{lstlisting}
You are a QA Verification Agent ({expert_persona}, {domain_context}). Evaluate the validity of the following QA pair.
Context: Users see the Question WITHOUT the Content. The Question must be completely self-contained.
Evaluation Rules:
1. **Standalone Principle (QUESTION_CORRECT | INCORRECT):**
   - The Question is INCORRECT if it relies on implicit context or vague references (e.g., "the provided figure", "this table", "the described method", "Section 2").
   - The Question is CORRECT only if it explicitly names the subject, standard, or artifact (e.g., "In Figure XX of the document YY", "The inflation-time chart...").
2. **Factuality (ANSWER_CORRECT | INCORRECT):**
   - The Answer must be factually supported by specific data/principles in the Content.
3. **Necessity (REQUIRES_CONTENT | CAN_ANSWER_WITHOUT_CONTENT):**
   - Determine if the specific provided Content is required to answer, or if general domain knowledge suffices.
Inputs:
Content: {content}
Question: {question}
Answer: {answer}
Required Output Format:
QUESTION_[CORRECT|INCORRECT]
ANSWER_[CORRECT|INCORRECT]
[REQUIRES_CONTENT|CAN_ANSWER_WITHOUT_CONTENT]
Justification: <Brief reasoning>
\end{lstlisting}

% ---------------------------------------------------------
\prompthead{PROMPTS["rerank_vlm"]}
\begin{lstlisting}
You are a VLM Reranking Agent. Rank the provided chunks (text and images) by relevance to the Query.
Input Markers: `<CHUNK_START id=N>` and `<IMAGE_START>`.
Instructions:
1. Analyze both textual context and visual data to determine relevance.
2. List ALL chunk IDs in descending order (Rank 1 = Highest).
3. Output strictly in this format (no conversational text):
<Rank 1>Chunk <id>
<Rank 2>Chunk <id>
<Rank N>Chunk <id>
\end{lstlisting}

% ---------------------------------------------------------
\prompthead{PROMPTS["deduplication_rank"]}
\begin{lstlisting}
You are a Data Curator ({expert_persona}, {domain}). Reorder the provided cluster of QA pairs based on their relationship to the core topic.
Task: Sort the list from **Most Distinct/Unique** (least similar) to **Most Representative/Redundant** (most similar).
Constraint: Preserve all text verbatim. Do not omit sub-questions or modify content.
Candidates:
{candidates_text}
Required Output Format:
<|#|>START<|#|>
Question<|#|><Question_Text><|#|>Answer<|#|><Answer_Text>
<|#|>NEXT<|#|>
Question<|#|><Question_Text><|#|>Answer<|#|><Answer_Text>
<|#|>END<|#|>
\end{lstlisting}

% ---------------------------------------------------------
\prompthead{PROMPTS["deduplication_merge"]}
\begin{lstlisting}
You are a Data Curator ({expert_persona}, {domain}). Synthesize the provided QA cluster into the MINIMAL set of high-quality pairs.
Processing Logic:
1. **Merge:** Combine complementary pairs into comprehensive questions (integrating sub-questions) and unified answers.
2. **Deduplicate:** Select the single best version for exact or near-duplicates.
3. **Goal:** Zero redundancy while retaining full information coverage.
Input Candidates:
{candidates_text}
Required Output Format:
<|#|>START<|#|>
Question<|#|><Merged/Refined Question><|#|>Answer<|#|><Merged/Refined Answer>
<|#|>NEXT<|#|>
Question<|#|><Second Pair (if needed)><|#|>Answer<|#|><Second Answer>
<|#|>END<|#|>
\end{lstlisting}

\end{document}